\documentclass{article}
\pdfoutput=1

\usepackage[preprint,nonatbib]{neurips_2020}

\usepackage[utf8]{inputenc} 
\usepackage[T1]{fontenc}    
\usepackage{url}            
\usepackage{booktabs}       
\usepackage{amsmath}        
\usepackage{amsfonts}       
\usepackage{nicefrac}       
\usepackage{microtype}      
\usepackage[pdftex]{graphicx}
\usepackage{subcaption}     
\usepackage{float}
\newtheorem{definition}{Definition}

\title{Concept-Based Explanations for Tabular Data}

\author{Varsha Pendyala\\
    varsha.pendyala@wisc.edu
   \And
   Jihye Choi\\
   jihye@cs.wisc.edu
}

\begin{document}

\maketitle
\begin{abstract}
The interpretability of machine learning models has been an essential area of research for the safe deployment of machine learning systems. 
One particular approach is to attribute model decisions to high-level concepts that humans can understand.
However, such concept-based explainability for Deep Neural Networks (DNNs) has been studied mostly on image domain. 
In this paper, we extend TCAV, the concept attribution approach, to tabular learning, by providing an idea on how to define concepts over tabular data.
On a synthetic dataset with ground-truth concept explanations and a real-world dataset, we show the validity of our method in generating interpretability results that match the human-level intuitions.
On top of this, we propose a notion of fairness based on TCAV that quantifies what layer of DNN has learned representations that lead to biased predictions of the model. Also, we empirically demonstrate the relation of TCAV-based fairness to a group fairness notion, Demographic Parity.
\end{abstract}

\section{Introduction}
\label{sec:intro}
Despite the successful generalizability of deep learning models in various application domains, limited understanding of their behavior prevents their deployment in full scale. The study of interpretability in machine learning gained greater attention recently, to explore what factors influence a model's prediction. In the real world, most commonly available and highly valuable data is in the form of tables. While the deep learning models are highly successful in computer vision, speech recognition and natural language processing, there is relatively less usage of such models for learning involving tabular datasets. As a result, majority of the work on interpretability of neural networks has not been studied in the context of tabular datasets.

Several approaches were proposed to explain the model's prediction to an input in terms of its input features. In case of deep neural networks for image understanding, feature attributions for model prediction are typically obtained as a saliency map over pixels of the input image \cite{smilkov1706smoothgrad, selvaraju2016grad, sundararajan2017axiomatic}. However, feature importance scores alone may not aid us in explaining the influence of high-level human friendly concepts on model predictions. For instance, how do we measure the influence of a pattern, object or the background in an input image on model's decision? 

Recent work shows the existence of detectors for higher level concepts at higher layers and lower level concepts at lower layers \cite{bau2017network}. This enables the network to converge to representations in a bottom-up manner while inducing linear separability of higher level features \cite{raghu2017svcca, alain2016understanding}. Supported by these findings, authors in \cite{kim2017interpretability} proposed a concept attribution approach: {\it Testing with Concept Activation Vectors} (TCAVs).

Through this project we extended the TCAV based interpretability to deep learning over tabular datasets. In addition, we show how the TCAV can be viewed as a tool for estimating model fairness. We considered a state-of-the-art neural network that is specially designed for the tabular datasets: {\it TabNet} \cite{arik2019tabnet}. Our experimental results show that TCAV based interpretability has been effective in revealing the biases present in the data used to train the TabNet.

\section{Related work}
\label{sec:related}
\subsection{Interpretability method based on concept attribution}
\label{sec:TCAV}
TCAV approach proposed in \cite{kim2017interpretability} generates insights by quantifying model's sensitivity with respect to user-defined concepts defined in the form of examples.

 \paragraph{Concept Activation Vector} Given a set of examples for every concept, the first step of TCAV is to find a vector for each concept in the space of activations at a layer {\it l}. 
 These vectors are referred as Concept Activation Vectors (CAVs). Each CAV points in the direction of activations of examples of corresponding concept. 
 CAV for a concept is normal to the hyperplane that separates positive and negative examples corresponding to that concept. 
 For instance, to learn a CAV for checkerboard pattern, positive examples can be all those images with checkerboard pattern and negative examples could be any random images which do not have that pattern. The hyperplane that separates them is its CAV.

\paragraph{Conceptual sensitivity}
Given a test image, it is possible to quantify the influence of a user-defined concept on the neural network's decision. 
For a concept $C$ with unit CAV vector $v_C^l$ computed from the activations of layer $l$ with $m$ nodes, sensitivity of class $k$ with respect to $C$ is given by:
\begin{align}
    S_{C,k,l}(x) &= \lim_{\epsilon \to 0}\frac{h_{l,k}(f_l(x)+\epsilon v_C^l) - h_{l,k}(f_l(x))}{\epsilon}\\
&= \nabla h_{l,k}(f_l(x))\cdot v_C^l \nonumber
\end{align}
where $f_l(x)$ denotes the activation of input $x$ computed at layer $l$ and $h_{l,k}:R^m \to R$.

TCAV score for any target class $k$ with respect to $C$ is computed using all the inputs belonging to the class $k$ as follows: when $|X_k|$ is the number of images from class $k$,
\[
TCAV_{C,k,l} = \frac{\sum_{x\in X_k} S_{C,k,l}(x)}{|X_k|}
\]

\subsection{Deep Neural Networks for Tabular data}
Unlike in the image domain, machine learning with canonical Deep Neural Networks (DNNs) remains under-explored, with variants of ensemble decision trees still dominating most applications. 
Tree-based methods have certain benefits indeed: they are representationally efficient, interpretable in their basic form, and fast to train. 
However, recent studies suggest that DNN-based architectures would lead to performance improvements, particularly for large datasets \cite{hestness2017deep}, and it is worth to explore deep learning for tabular data.
Since canonical DNNs based on stacked convolutional layers or multi-layer perceptrons are likely to be overparametrized over tabular data \cite{GoodBengCour16}, the need for a DNN architecture suitable for tabular learning emerges.

\begin{figure}[H]
    \centering
    \includegraphics[width=1.0\textwidth]{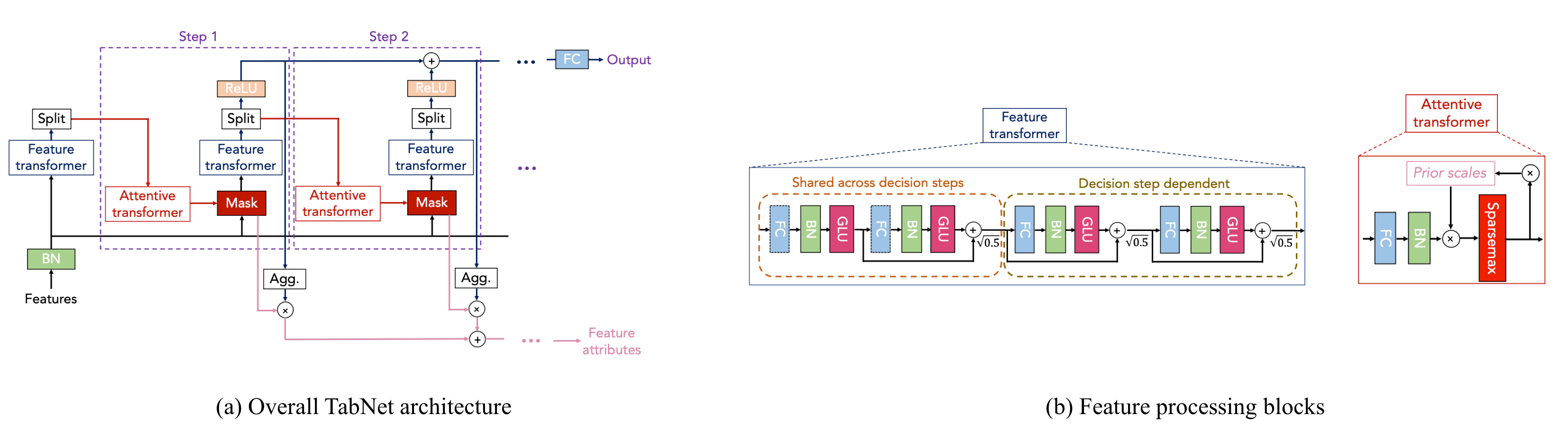}
    \caption{TabNet architecture from \cite{arik2019tabnet}. (a) Using a sequential attention mechanism, TabNet chooses a subset of semantically meaningful features to process in each decision step. A complete feature vector $X$ is passed to each decision step. (b) $i^{th}$ step uses the processed information from $(i-1)^{th}$ step to process subset of features of $X$ using feature transformers. Each step outputs the processed feature representation to be aggregated into the overall decision.}
    \label{fig:tabnet}
\end{figure}

\paragraph{TabNet}
Arik \textit{et al.} \cite{arik2019tabnet} proposed a novel DNN architecture that uses sequential attention to choose which features to reason from at each decision step (Figure \ref{fig:tabnet}).
TabNet outperforms other neural network and decision tree variants across tabular datasets from diverse domains.
Throughout this paper, we adopt the TabNet architecture for classification tasks. 

\section{TCAV based interpretability for tabular data}
\label{sec:tcav-tabnet}
\subsection{User-defined concepts for tabular data}
We apply the TCAV approach to explain the influence of user-defined concepts over classifications tasks defined on tabular datasets. Given that the tabular data is structured, we define concepts over columns of table data in the form of boolean predicates. This automates the process of providing positive and negative examples to the binary classifier learning a CAV.

\begin{figure}[H]
    \centering
    \includegraphics[width=0.45\textwidth]{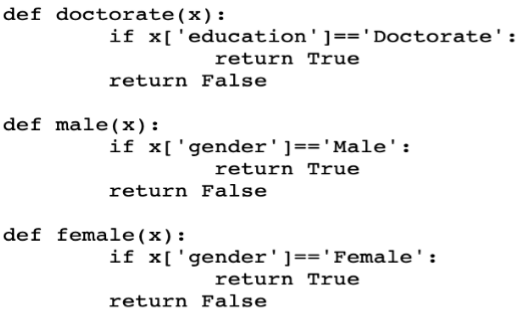}
    \caption{Predicates over Adult Income Dataset}
    \label{fig:predicate}
\end{figure}

Example concepts for Adult Income Dataset \cite{Dua:2019} are shown in Figure \ref{fig:predicate}.
\subsection{Training CAVs with tabular data}
We train {\it CTGAN}, a generative model described in \cite{xu2019modeling} to generate satisfying and unsatisfying assignments to our boolean predicates for all user-defined concepts. Through our experiments in section \ref{subsec:tcav-AID} we show that CTGAN has been very effective in computing reliable CAVs. 

Now we describe how using a CTGAN to generate data for concept learning in case of non-pictorial data is equivalent to manually providing positive, negative example images for concept learning in case of image domain.

\paragraph{Concepts in image domain}Let $stripes$ denote the user-defined concept. All the images with stripes can be thought of samples drawn from a distribution $Stripe_X$. Note that the positive images for $stripes$ concept are not sampled from the original data distribution $D_{XY}$ where $X$ is the random vector corresponding to all the pixels of an input image, $Y$ is the class label.
$$\text{striped images} \sim Stripe_X$$
If we know the distribution $Stripe_X$, all the positive images for $stripes$ concept can be drawn from it and given to the linear classifier to obtain the CAV vector. However, since we do not know $Stripe_X$, we would create synthetic images with stripes to feed to the binary linear classifier. Since it is pictorial data, it is possible for us to assign intensity values to different pixels to compose realistic striped images that captures joint distribution $Stripe_X$ among its pixels. 

\paragraph{Concepts in tabular domain}Let \textit{age $<$ 25} denote the user-defined concept. Denote the original data distribution as $D_{XY}$, where $X$ denotes the random vector corresponding to all the columns of an input row, $Y$ is the class label. Assume $A_X$ to be the distribution from which positive examples for \textit{age $<$ 25} can be drawn and $B_X$ from which negative examples can be drawn.
$$\text{input data with age $<$ 25} \sim A_X$$
$$\text{input data with age $\geq$ 25} \sim B_X$$
Since we do not know $A_X, B_X$, we can provide examples of row data with age $<$ 25, age $\geq$25 to the binary classifier to learn CAV. Since it is non-pictorial data, it may not be possible for us to assign values to different column features to compose realistic age $<$ 25 or age $\geq$ 25 data that captures respective join distributions $A_X, B_X$ among its columns. 

So, we can follow below procedures to generate examples that captures joint distributions $A_X, B_X$.
\paragraph{Procedure 1}
\begin{itemize}
    \item Use the available natural data (i.e train/test data) where age $<$ 25 to train a $CTGAN_{age < 25}$.
    \item Use the natural data where age $\geq$ 25 to train $CTGAN_{age \geq 25}$.
    
    Note that sampling the data from $CTGAN_{age < 25}$ or $CTGAN_{age \geq 25}$ is equivalent to sampling from $A_X$ or $B_X$ respectively.
    \item Sample examples from $CTGAN_{age < 25}$, $CTGAN_{age \geq 25}$ to learn the binary classifier corresponding to CAV. 
\end{itemize}
For a random vector $r$ representing a table row, we have:
$$Pr_{A_X}(r) = Pr_{D_X}(r|r['age']<25)$$
$$Pr_{B_X}(r) = Pr_{D_X}(r|r['age']\geq25)$$
Using the above relations, we can follow the below simplified procedure to generate examples which capture $A_X, B_X$ joint distributions:
\paragraph{Procedure 2}
\begin{itemize}
    \item Use the entire available natural data (i.e train/test data) to train a $CTGAN_X$.
    
    Note that sampling the data from $CTGAN_X$ is equivalent to sampling from $D_X$.
    \item Then sample the datapoints  from  $CTGAN_X$  to  accumulate  positive  and  negative  examples  for  every  concept  we  are interested in learning.
\end{itemize}

\section{TCAV based fairness}
\label{sec:tcav-fairness}
This section describes our idea and method: how to use TCAV based interpretability for tabular data (Section \ref{sec:tcav-tabnet}) as a tool for diagnosing fairness level of a trained models. 

Without loss of generality, we consider a classifier $f: \mathbb{R}^n \rightarrow \mathbb{R}^m$ with inputs $X \in \mathbb{R}^n$ and a output layer with $m$ neurons. 
Let $Y$ be a target label such that $Y = \{y_1, y_2, ..., y_m\}$ and $A = {A_1, A_2, A_3, ...}$ be a protected features among $n$ features of X (e.g. A = {'female', 'male'}, for a protected feature 'gender').
We assume $X, A, Y$ are generated from an underlying distribution $D$ i.e. $(X, A, Y) \sim D$.

The first step in our method is to define a concept of interest $C$ in terms of $A$. 
More specifically, a concept is described as a predicate that assigns true for samples having $A_1$ (that is, satisfying examples) and false for samples without $A_1$ (that is, unsatisfying examples).
Then, by setting target class as $y_k \in Y$, we train two linear classifiers to obtain CAV vector $v_{C_1}^l$ and $v_{C_2}^l$ in Equ. \ref{eqn:concept-trend}, where $C_1$ is $X[A] == A1$ and $C_2$ is $X[A] != A1$, respectively. 

Given the above, we propose a definition of fairness that involves TCAV scores for $C_1, C_2, y_k$ \ref{def:tcav-fairness}. The intuition behind the definition is that if a classifier's prediction is independent to the protected attribute, its conceptual sensitivity gauged over samples that possess that attribute or not, should be equal.
\begin{definition}
  \label{def:tcav-fairness}
  A classifier $f$ satisfies fairness in regards to concept attributions under a distribution $D$ if its prediction $f(X)$ is equally sensitive to $C_1$ and $C_2$ - that is, if $|TCAV_{C_1,k,l} - TCAV_{C_2,k,l}| = 0$ for all $y_k \in Y$.
\end{definition}

Considering that TCAV score is basically a dot product between two vectors, $h_{l,k}(f_l(x))$ and $v_C^l$, we introduce to quantify the sensitivity of a prediction to a concept as an angle between the two vectors. 
\begin{equation}
    \label{eq:tcav-angle}
    \angle \: TCAV_{C,k,l} = \frac{\sum_{x\in X_k} \angle \: (h_{l,k}(f_l(x)), v_C^l)}{|X_k|}
\end{equation}
where $\angle \: (h_{l,k}(f_l(x)), v_C^l) = \cos^{-1}(\frac{h_{l,k}(f_l(x))\cdot v_C^l)}{||h_{l,k}(f_l(x))|| ||v_C^l)||})$. Using the notion of TCAV angle, as the angle between $h_{l,k}(f_l(x))$ and $v_C^l$ gets close to $90^\circ$, the classifier becomes fairer. Our definition of fairness in \ref{def:tcav-fairness} can be rewritten as follows.
\begin{definition}
  \label{def:tcav-fairness-angle}
  A classifier $f$ satisfies fairness in regards to concept attributions under a distribution $D$ if $\angle \: TCAV_{C,k,l} = 90^\circ$ for all $y_k \in Y$ and $C = \{C_1, C_2\}$.
\end{definition}

The advantage of our definition over other fairness definitions such as Demographic Parity or Equalized Odds \cite{hardt2016equality} is that it allows to tell what layers of neural networks have learned representations that contribute to biased predictions.

\section{Experiments}
\label{sec:experiments}

Through our experiments we provide empirical evidence to the following: {\bf (1)} TCAV scores of a target class with respective to various user-defined concepts can throw light on the inherent biases within the training data. {\bf (2)} Generative models can be very effective in generating example data for training reliable CAVs. {\bf(3)} TCAV can be viewed as a fairness estimation tool.

We perform experiments on classification tasks in synthetic settings and in published benchmark settings.
We show evidence that TCAV results align with intended concepts of interest, with tabular datasets that are also studied in \cite{arik2019tabnet}. 
We use the same input data processing, and hyperparameters for training TabNets as in \cite{arik2019tabnet}.

Based on the open-source implementations \cite{tcav-code, tabnet-code}
we develop codes for obtaining TCAV interpretability on tabular datasets in Tensorflow.
For every user-defined concept, a CAV is trained given the activation layer outputs that correspond to the tensors resulting from the cumulative sum over ReLU outputs at each decision step in Figure \ref{fig:tabnet}(a). 

\subsection{TCAV based interpretability}\label{subsec:tcav-AID}

\paragraph{Synthetic Dataset} We consider a synthetic tabular dataset for binary classification task from \cite{chen2018learning} consisting 1M training samples and a separate test set of 100k samples.
The synthetic dataset is constructed in such a way that the ground truth output of the dataset $Y$ only depends on the two dimensional XOR of first two features $X1$ and $X2$.
Hence, only $X1$ and $X2$ serve as true features, where each dimension itself is independent of $Y$ but the combination of them has a joint effect on $Y$.
More specifically, for a sample to belong to the class of $Y = 0$, it is implied to have $X1X2 > 0$, whereas the output label $Y$ is 1, if and only if the sample has the property of $X1X2 < 0$.

\begin{figure}
    \centering
    \includegraphics[width=0.6\textwidth]{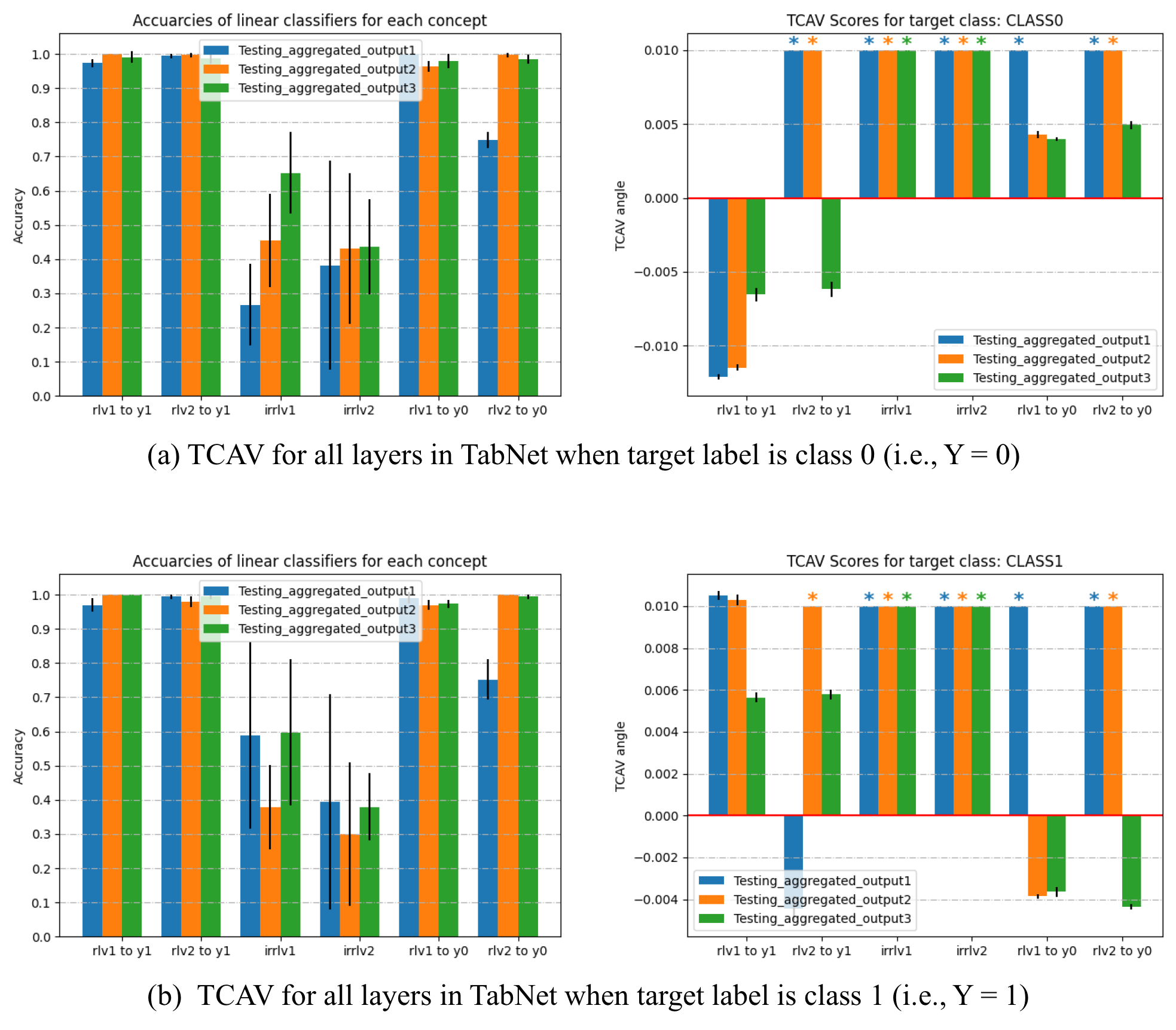}
    \caption{The accuracies of CAVs and TCAV scores over synthetic dataset. On the X axis, 'rlv1 to y1', 'rlvl to y1', 'rlv1 to y0' and 'rlv1 to y0' correspond to C1, C2, C5 and C6, respectively. 'irrlv1' and 'irrlv2' represent C3 and C4. '\text{*}'s mark CAVs ommitted after statistical testing.}
    \label{fig:synthetic}
\end{figure}

The concepts of our interest for each of which CAVs are trained are as follows: 
\begin{itemize}
    \item C1: $X1 > 0$ AND $X2 < 0$
    \item C2: $X1 < 0$ AND $X2 > 0$
    \item C3: $X5 > 0$
    \item C4: $X10 < 0$
    \item C5: $X1 > 0$ AND $X2 > 0$
    \item C6: $X1 < 0$ AND $X2 < 0$
\end{itemize}
When $Y = 1$, C1 and C2 are the positively relevant concepts, where C5 and C6 are negatively relevant concepts to the predictions.
On the contrary, when $Y = 0$, having C1 and C2 affects the output negatively, but C5 and C6 become positively relevant concepts. 
For both of the cases, C3 and C4 are irrelevant concepts to determining the output label.
We observe that the empirical results align with this intended design of concepts. 
In Figure \ref{fig:synthetic}(a), TCAV scores for C1 and C2 are negative values, whereas they are positively large in Figure \ref{fig:synthetic}(b). 
It means that having properties defined as C1 and C2 satisfied leads to a high chance for the sample to be predicted as class 1. 
This applies to C5 and C6 in the opposite manner and agrees with our intuition as well.
Please note that the accuracies of linear classifier for C3 and C4 are no better than random guessing, and resulting CAVs are not meaningful enough, so ommitted after statistical testing as proposed in \cite{kim2017interpretability}.

\paragraph{Adult Income Dataset \cite{Dua:2019}} 
We consider Adult Income Dataset, which is the most commonly used benchmark dataset for binary classification. 
The task is to distinguish whether a persons income is above \$50k from demographic and employment related features. 
CAVs are trained for following five concepts:
\begin{itemize}
    \item {C1}: age < 25
    \item {C2}: native-country == Mexico
    \item {C3}: gender == Female
    \item {C4}: native-country == India
    \item {C5}: education == Doctorate.
\end{itemize}

In the training data, we observed following trend with respect to the above mentioned sub-groups:
\begin{equation}\label{eqn:concept-trend}
Pr(Y=0|C1) \geq Pr(Y=0|C2) \geq Pr(Y=0|C3) \geq Pr(Y=0|C4) \geq Pr(Y=0|C5)
\end{equation}
where $Y=0$ for an individual implies, their income is less than $50K$.

Table \ref{table:cav-train} lists four different ways in which we trained CAVs and computed TCAV scores.
\begin{table}
  \caption{Datasets used to obtain concept attribution. 
  Dataset to train CAVs is a set of data from which the satisfying and unsatisfying assignments of given predicates are taken to train the CAVs.
  Dataset to compute TCAV score is a set of data from which the inputs corresponding to the given target class are used to compute TCAV score.
}
  \label{table:cav-train}
  \centering
  \begin{tabular}{lll}
    \toprule
    { Dataset to train CAVs}     & { Dataset to compute TCAV score}  \\
    \midrule
    Training data of TabNet & Test data \\
    Test data     & Test data & \\
    Synthetic data     & Test data\\
    GAN data & Test data\\
    \bottomrule
  \end{tabular}
\end{table}

 We generated synthetic data mentioned in Table \ref{table:cav-train} by uniformly sampling each column of the table in it’s respective domain: numerical features are sampled from the uniform distribution with $\min$ and $\max$ corresponding to the respective numerical column in training data, categorical data is uniformly sampled from all the unique values corresponding to the respective categorical column in training data.  
 
\begin{figure}
    \centering
    \includegraphics[width=1\textwidth]{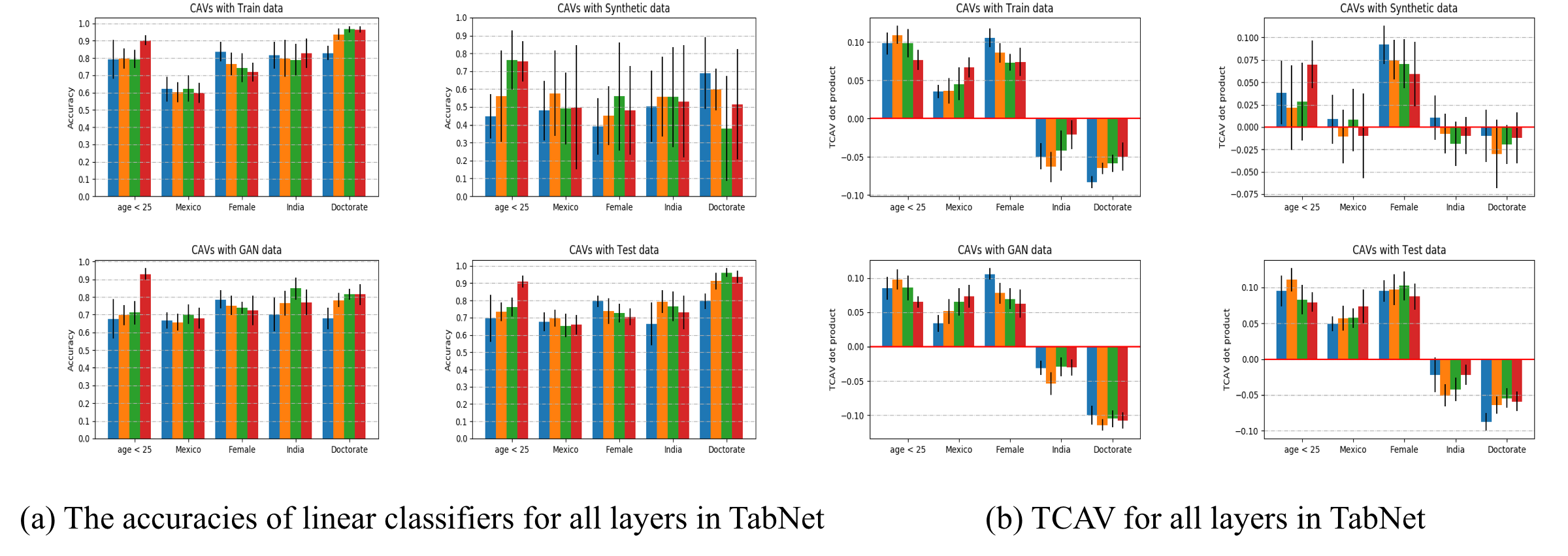}
    \caption{The accuracies of CAVs and TCAV scores over Adult Income Dataset}
    \label{fig:tcav_four}
\end{figure}
 \begin{figure}
    \centering
    \includegraphics[width=0.5\textwidth]{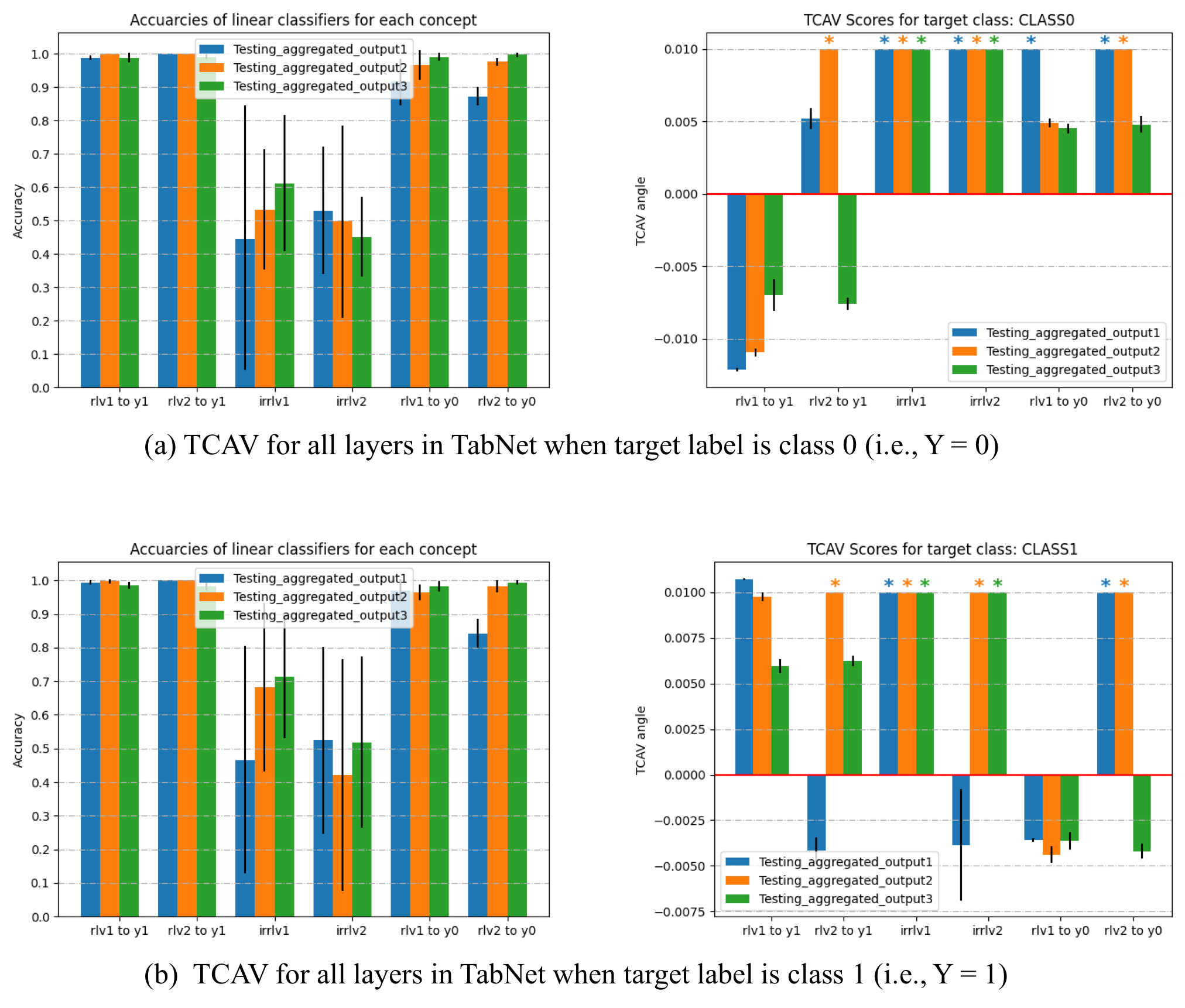}
    \caption{The accuracies of CAVs and TCAV scores over synthetic dataset. The only difference with the settings of Figure \ref{fig:synthetic} is that the values of features corresponding the concepts are replaced with values uniformly drawn within the range of the original values}
    \label{fig:synthetic-uniform}
\end{figure}

 From the results shown in Figure \ref{fig:tcav_four}(b) we can see that the TCAV scores across user-defined concepts indeed reveal the biases within the training data by following the same trend across concepts as in \ref{eqn:concept-trend}. Different colors in figures \ref{fig:tcav_four}(a), \ref{fig:tcav_four}(b) correspond to different decision step activations: $blue$ to $red$ for first layer to the last one.
 
 We can see that the GAN based TCAV scores in Figure \ref{fig:tcav_four}(b) with various user-defined concepts follow the same trend as those for which CAVs were trained with training or test data. This implies, when we do not have enough test data available for training CAVs, we can use GAN to generate satisfying and unsatisfying assignments for the boolean predicates.
 
 Also we observed that the assignments coming from the synthetic data could not result in reliable CAVs. This can be noted from the very high standard deviation of the accuracy of linear classifiers at every decision step, when synthetic data is used to train the CAVs (shown in Figure \ref{fig:tcav_four}(a)). 

To further verify our argument on why the interpretability fails with the modified data, we perform the same experiment in the synthetic setting that is described early in this section. 
Unlike in the real-world data, the features in domain $X$ are independent of each other, so that we expect to obtain reliable CAVs even with uniformly sampled data.
Figure \ref{fig:synthetic-uniform} shows almost identical interpretability results with \ref{fig:synthetic}.

\subsection{TCAV based fairness}
To show the validity of our fairness definition \ref{def:tcav-fairness}, \ref{def:tcav-fairness-angle}, we empirically demonstrate that the fairness level quantified by our definition aligns with that of other commonly used fairness metrics.
Since TCAV provides global interpretability that is measured not on a single samples, but all samples in a class, it is natural to consider the TCAV-based fairness notions as a global fairness. 
In this section, we investigate the correspondence between TCAV-based fairness and a famous group fairness metric, Demographic Parity.

Demographic Parity can be thought of as a version of the US Equal Employment Opportunity Commission’s “four-fifths rule” that requires “selection rate for any race, sex, or
ethnic group [must be at least] four-fifths (4/5) (or eighty
percent) of the rate for the group with the highest rate.”
In our experiments, we use Adult Income Dataset where $Y$ = \{'>50K', '<=50K\} and a binarized protected feature 'gender', i.e. $A$ = \{'female', 'male'\}. CAVs are trained for two concepts, C1: X['gender'] == 'female' and C2 : X['gender'] == 'male'. Then, Demographic Parity can be quantified as follows,
\[\epsilon = Pr[\hat{y} = '>50K' | A = 'Female'] - Pr[\hat{y} = '>50K' | A = 'Female']\]
where $\hat{y}$ is the predicted label by a trained model. As $\epsilon$ is close to zero, the classifier is fair.

 \begin{figure}
    \centering
    \includegraphics[width=0.8\textwidth]{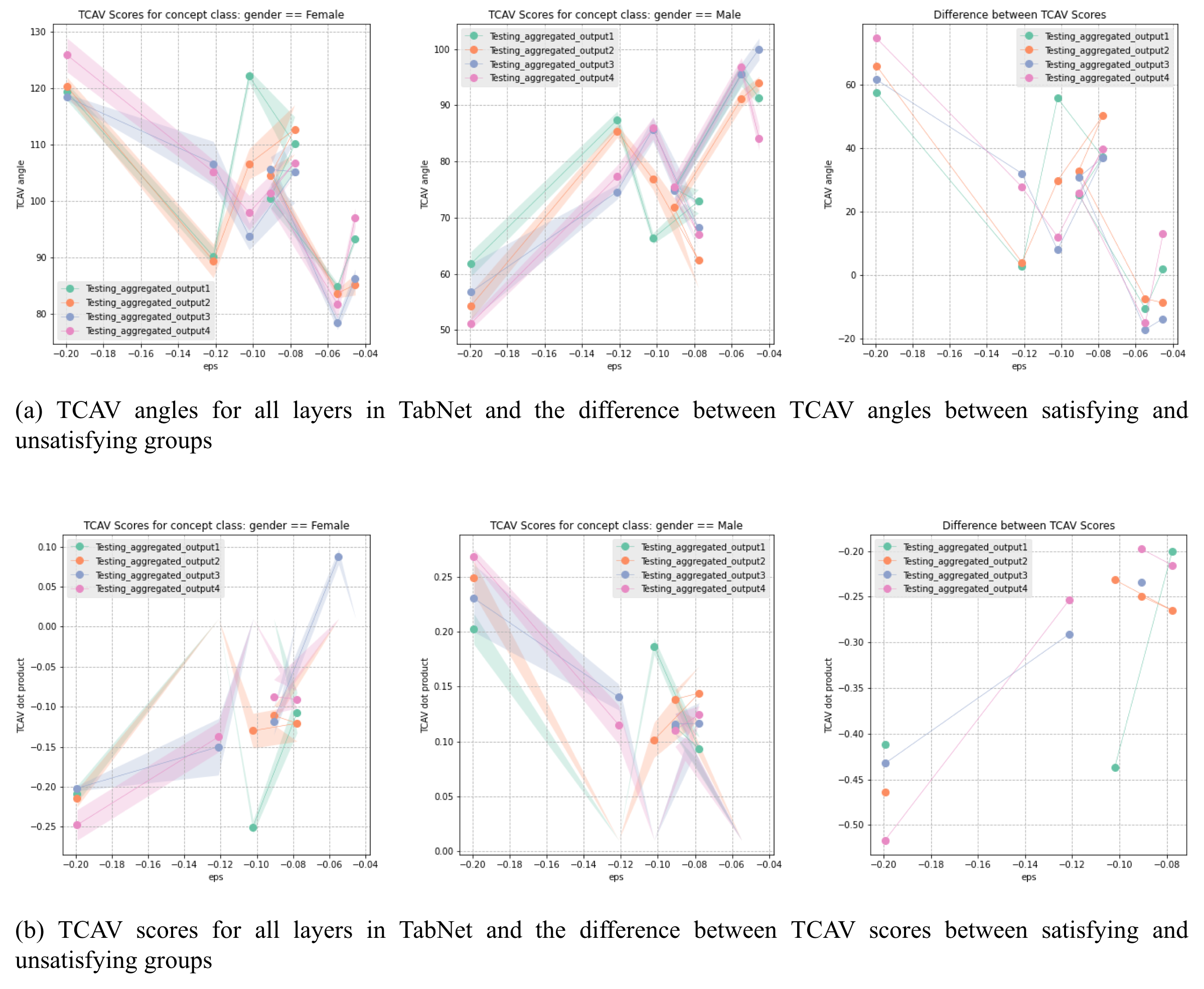}
    \caption{TCAV resuls given a binarized protected feature 'gender'}
    \label{fig:AID-fairness}
\end{figure}

With multiple versions of modified training data that are unbalanced in terms of $\epsilon$, TabNet models are trained to have different fairness levels of Demographic Parity. 
Figure \ref{fig:AID-fairness} shows TCAV scores and TCAV angles of multiple TabNets. 
By the definition, when $\epsilon$ is far from zero in negative direction, it implies that the classifier is more biased toward $A$ = 'male', and we observe that it is agreed with TCAV results: in Figure \ref{fig:AID-fairness}(a), TCAV angle is much larger for C1 but is smaller for C2 compared to $90^\circ$, and in Figure \ref{fig:AID-fairness}(b), TCAV score is negatively large for C1, but positively large for C2. For both of the plots, the difference between TCAV results gets close to zero as $\epsilon$ goes to zero. 

\section{Conclusion and future work}
\label{sec:conclusion}
In this paper, we presented a method to extend the use of concept attribution interpretability, TCAV, to the domain of tabular learning using TabNet. 
Our experiments suggested TCAV results in interpretability that agrees with natural high-level concepts indeed.
We then showed how they can be used to quantitatively measure the fairness notion of a trained neural networks, and provided empirical evidence on the correspondence between TCAV-based fairness and group fairness such as Demographic Parity.

TCAV can be used to any deep neural networks where we can extract deep representations from layers of the model. Hence, the fairness tool presented here is applicable to other deep neural networks for tabular learning, other than TabNets.
Moreover, there are several promising avenues for future work based on our TCAV-based fairness. 
While we have provided empirical evidence on its relation to Demographic Parity, studying its correspondence to other group fairness notions or providing theoretical evidence would bring new insights.
Finally, one could ask for ways to extend this idea for further sophisticated diagnosis of fairness, such as finding a proxy for a protected features.

\section{Acknowledgement}
This work was done as a part of the course CS 839: Verified Deep Learning (Spring 2020) offered by Prof. Aws Albarghouti. The support from Prof. Albarghouti and the course staff is gratefully acknowledged.

\small
\bibliographystyle{ieeetr}
\bibliography{references}
\end{document}